\documentclass[lettersize,journal]{IEEEtran}
\usepackage{amsmath,amssymb,amsfonts}
\usepackage{algorithmic}
\usepackage[ruled]{algorithm2e}
\usepackage{array}
\usepackage{textcomp}
\usepackage{stfloats}
\usepackage{url}
\usepackage{color}
\usepackage{verbatim}
\usepackage{graphicx}
\usepackage{multirow}
\usepackage{array}
\usepackage{booktabs}
\usepackage{graphicx}
\usepackage{parskip}

\usepackage[numbers]{natbib}
\usepackage{bm}
\usepackage{mathtools}

\usepackage{subfigure}
\usepackage{makecell}

\usepackage{bibspacing}
\graphicspath{ {./fig/} }

\usepackage{flushend}

\usepackage{stackengine}

\def\BibTeX{{\rm B\kern-.05em{\sc i\kern-.025em b}\kern-.08em
    T\kern-.1667em\lower.7ex\hbox{E}\kern-.125emX}}
\usepackage{balance}
\begin{document}

\title{Chain-of-Trust: A Progressive Trust Evaluation Framework Enabled by Generative AI} 
\author{Botao~Zhu, Xianbin~Wang, Lei~Zhang, and  Xuemin (Sherman) Shen
\thanks{
B. Zhu and X. Wang are with the Department of Electrical and Computer Engineering, Western University, London, Canada N6A 5B9 (Emails: \{bzhu88, xianbin.wang\}@uwo.ca)

L. Zhang is with the James Watt School of Engineering, University of Glasgow, Glasgow,
U.K. G12 8QQ (Email: Lei.Zhang@glasgow.ac.uk)

X. Shen is with the Department of Electrical and Computer Engineering, University of Waterloo, Waterloo, Canada N2L 3G1 (Email: sshen@uwaterloo.ca)}}


\maketitle

\begin{abstract}

In collaborative systems with complex tasks relying on distributed resources, trust evaluation of potential collaborators has emerged as an effective mechanism for task completion. However, due to the network dynamics and varying information gathering latencies, it is extremely challenging to observe and collect all trust attributes of a collaborating device concurrently for a comprehensive trust assessment. In this paper, a novel progressive trust evaluation framework, namely chain-of-trust, is proposed to make better use of misaligned device attribute data. This framework, designed for effective task completion, divides the trust evaluation process into multiple chained stages based on task decomposition. At each stage, based on the task completion process, the framework only gathers the latest device attribute data relevant to that stage—leading to reduced trust evaluation complexity and overhead. By leveraging advanced in-context learning, few-shot learning, and reasoning capabilities, generative AI is then employed to analyze and interpret the collected data to produce correct evaluation results quickly. Only devices deemed trustworthy at this stage proceed to the next round of trust evaluation. The framework ultimately determines devices that remain trustworthy across all stages. Experimental results demonstrate that the proposed framework achieves high accuracy in trust evaluation.


\end{abstract}


\section{Introduction}

\IEEEPARstart{W}{ith} the increasing complexity of applications and interconnected systems, it becomes impractical for individual devices—with their limited computational capacity and energy supplies—to handle computing tasks on their own~\cite{9964025}. To address this challenge, researchers have started exploring the collaborative use of distributed resources to execute tasks cost-effectively through distributed computing~\cite{9026979},~\cite{1122}. This approach harnesses the collective capabilities of multiple collaborating devices to tackle tasks that would overwhelm a single device~\cite{10239356}. Currently, device collaboration-based systems are being widely applied across various industries, such as manufacturing, smart cities, and e-health, to provide low-latency and high-quality services~\cite{10388361}.

\begin{figure}[!t]
\centering
\includegraphics[scale=0.98]{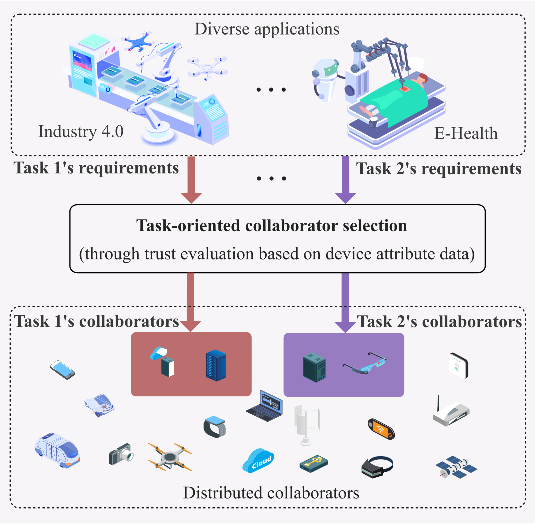}
\caption{A task-oriented collaborator evaluation and selection mechanism in collaborative systems with diverse applications and distributed collaborators.}
\label{6G_system}
\end{figure}

To ensure the reliable completion of collaborative tasks, the evaluation and selection of collaborators have become a central issue in collaborative systems. {With the increasing complexity of tasks, an effective collaborator evaluation mechanism must account for both well-established factors—such as security, privacy, and reliability—and task-specific requirements, including capabilities and available resources. Recent studies have proposed using trust as a means to quantify collaborator reliability~\cite{10109152},~\cite{Botao},~\cite{10229502},~\cite{10734666}. In these approaches, trust is typically defined as the task owner's expectation of a collaborator’s capability, inferred from various device attributes such as service types, behavioral history, performance records, and resource availability. However, this general-purpose evaluation is insufficient for collaborative environments characterized by heterogeneous applications and distributed resources. In such settings, trust should be assessed in a task-oriented manner, reflecting the collaborator’s ability to fulfill specific task requirements.}
As shown in Fig.~\ref{6G_system}, based on diverse task requirements at the application layer, relevant device attribute data from distributed collaborators is collected to accurately evaluate and select the trusted collaborators. However, there are several challenges that need to be addressed to implement this mechanism.



First of all, all device attribute data may not be observed and collected simultaneously due to factors such as network delays, resource limitations, or asynchronous updates. Additionally, attempting to collect all device attribute data at once demands substantial resources, which increases the overall system burden and evaluation complexity. Therefore, developing a low-complexity, progressive trust evaluation approach that gradually acquires device attribute data presents a practical and efficient solution. 




Moreover, accurately understanding complex task requirements remains a key challenge in collaborative systems. Tasks often vary in their demands for computational resources, communication capabilities, and service quality. As a result, assessing the trustworthiness of devices requires evaluating their ability to fulfill specific task requirements rather than relying on generalized metrics. Traditional trust evaluation approaches struggle to adapt to these dynamic conditions, leading to potential mismatches between task requirements and device capabilities. Therefore, using technologies like artificial intelligence (AI) to enable intelligent adaptation and understanding of varying task requirements is essential in trust evaluation.



Furthermore, accurately understanding the collected device attribute data and capturing changes in these data, such as performance, behavior, and available resources, requires careful consideration. The performance and behavior of devices reflect their past reliability, while their available resources represent their ability to perform future tasks. These factors collectively determine the devices' trustworthiness. Thus, the trust evaluation approach needs to utilize AI technologies to capture and comprehend these factors, ensuring that the capabilities of devices consistently align with task requirements.

To address the challenges mentioned above, this paper proposes a novel, progressive trust evaluation framework—chain-of-trust. Instead of collecting all device attribute data simultaneously, chain-of-trust divides trust evaluation into multiple stages—task requirement decomposition, service availability evaluation, communication resource evaluation, computing resource evaluation, and result delivery evaluation—and gradually collects device attribute data at each stage. Device trustworthiness is evaluated sequentially throughout these stages, where each evaluation result depends on the outcome of the previous stage. As trust evaluation involves complex reasoning over multiple factors, generative AI—known for its strength in such tasks—is leveraged to implement the proposed chain-of-trust. The main contributions of this paper are summarized as follows: 

\textbullet \, We introduce chain-of-trust in collaborative systems for the first time. It divides the trust evaluation process into multiple stages. At each stage, the system efficiently utilizes a small amount of resources to rapidly collect the latest device attribute data based on task requirements and assesses the trustworthiness of devices in meeting those requirements. Only devices deemed trustworthy at the current stage advance to the next evaluation, allowing for gradual refinement and validation of trustworthiness.

\textbullet \, Generative AI-enabled chain-of-trust provides an accurate understanding of task requirements and the collected device attribute data at each stage, ensuring precise trust evaluation results. Additionally, it relies on few-shot prompting, allowing it to dynamically adapt to new tasks without training.

\section{Limitations of Existing Research on Trust-Based Collaboration}


Existing research on trust in collaborative systems mainly suffers from the following limitations that need to be addressed.

    \textbullet \, \textbf{Viewing trust as a synonym for security}. Most studies equate trust with security, relying solely on security-related technologies, such as authentication and blockchain, to ensure the trustworthiness of collaborators and the security of information transmission~\cite{10470435},~\cite{10001120}. In fact, security should be considered a part of trust and is one of the services that devices can provide. The factors determining trust should be more comprehensive, encompassing all aspects of devices related to ensuring task completion, such as communication resources, computing resources, and energy.

    \textbullet \, \textbf{Overlooking that trust is task-specific}. 
    Most studies typically assume that trust relationships between devices are one-to-one, evaluating them by collecting data such as historical behavior and feedback. However, this evaluation approach fails to accurately capture the true trust levels between devices. In reality, trust between devices should be task-specific, meaning the same device may demonstrate varying levels of trustworthiness depending on different task requirements. Moreover, the trustworthiness of devices can change with variations in their attributes, such as available resources.


    \textbullet \, \textbf{Ignoring the fact that device attribute data may not be updated simultaneously}. Most studies assume that all device attribute data is available before the start of trust evaluation. However, {in practical systems, factors such as network latency and instability, heterogeneous device processing capabilities, and unsynchronized system clocks result in inconsistent update frequencies of device attributes, which makes it difficult to obtain all the required up-to-date information simultaneously.} This incompleteness directly affects the accuracy and comprehensiveness of trust evaluation.
    
   {In response to the aforementioned limitations, the design of a progressive trust evaluation scheme—capable of stage-wise collection and evaluation of asynchronous data—offers a compelling approach to achieving accurate and task-specific trust evaluation.}

\section{Advantages of Generative AI}
The advanced technological strengths of large language models (LLMs)—including exceptional versatility, powerful generative capabilities, and wide adaptability—have allowed generative AI to excel in addressing a broad range of complex tasks~\cite{10795202},~\cite{10648594}. Since trust evaluation is a complex reasoning task involving multiple factors, it is well within the capabilities of generative AI. The main advantages of generative AI and an analysis of how they can solve specific technical challenges in trust evaluation are outlined below. 

\textbullet \, \textbf{In-context learning}. Generative AI exhibits advanced contextual understanding, allowing them to navigate complex scenarios, interpret nuanced input meanings, and produce coherent, high-quality content~\cite{10483549}. Furthermore, their ability to dynamically adjust outputs based on user requirements ensures adaptability across diverse contexts. In trust evaluation, this ability can provide in-depth analysis and understanding of complex factors such as task requirements, services and resources, providing a reliable basis for trust assessment.

\textbullet \, \textbf{Zero-shot learning and few-shot learning}.
Generative AI leverages a vast pre-trained knowledge base to perform new tasks without task-specific training (zero-shot learning) or uses a small number of examples (few-shot learning) to rapidly adapt to new task requirements~\cite{10746594}, reducing dependency on large datasets and lowering development barriers. In trust evaluation, when factors such as task requirements or resources change, these capabilities allow generative AI to rapidly adapt, facilitating efficient and dynamic trust evaluation with minimal or no additional training.

\textbullet \, \textbf{Chain-of-thought reasoning}. 
Generative AI enables step-by-step reasoning by decomposing complex tasks into a sequence of logical steps that are processed to derive solutions~\cite{10746594}.
When applied to trust evaluation, it enables the process to be deconstructed into clear, incremental stages, ultimately improving both accuracy and reliability through gradual analysis and reasoning.

{
\section{Chain-of-Trust}
\label{task_completion}


To support accurate trust evaluation for effective task completion in collaborative systems with constrained resources and asynchronous device attribute data, this section introduces the chain-of-trust framework. The framework is implemented on a central server, which is assumed to operate honestly. The server handles the collection of device attribute data based on the task owner's requirements, performs trust evaluations, and oversees the entire collaboration process. Chain-of-trust decomposes the trust evaluation process into multiple stages, beginning with task requirement decomposition and progressively achieving service availability evaluation, communication resource evaluation, computing resource evaluation, and result delivery evaluation. At each stage, only the latest device attribute data relevant to that stage is collected to assess device trustworthiness, and only devices meeting the trust criteria proceed to the next stage. This staged approach reduces overall evaluation complexity and allows resources to be allocated as needed at each stage, avoiding the high cost of acquiring all device attribute data at once. Owing to the advantages of generative AI, we employ it to implement the chain-of-trust framework, as shown in Fig.~\ref{flowchart}, enabling step-by-step and dynamic trust assessment.


In this section, we first introduce a real-world collaborative system, and then detail the implementation of the chain-of-trust framework. 

\begin{figure}[!]
\centering
\includegraphics[scale=0.91]{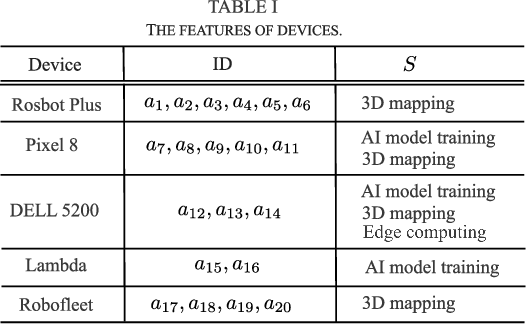}
\label{table1}
\end{figure}

\subsection{Collaborative system model}

 A collaborative system is constructed, comprising five Google Pixel 8 phones, three DELL 5200 servers, six Rosbot Plus robots, four Robofleet robots, and two Lambda GPU workstations. The complete device set is denoted as $A = \{a_1, a_2, \dots, a_{20}\}$, as shown in Table I. In addition, a different DELL 5200 is deployed as the control server, responsible for collecting data based on task owner's requests, executing the chain-of-trust framework, and monitoring the entire collaboration process. Each device can act as a task owner, generating a task and seeking assistance from the control server to select trusted collaborators. We assume that $a_1$ generates a task $T$ at time $t$,  which includes specific task information as well as the relevant metrics $a_1$ aims to achieve, such as quality of service (QoS), security, and quality of experience (QoE). For example, if $T = $ \textit{``I have several photos of an environment and want to quickly and securely accomplish a 3D mapping task based on these photos, which devices can be trusted to perform this task?''}, the phrase \textit{``a 3D mapping task''} specifies the service, \textit{``quickly and securely''} denotes the QoS, QoE, and security requirements.
For any device $a_m \in A$,  the set of services and resources it can provide is represented as $a_m = \{S, C^{\text{rate}}, C^{\text{sec}}, P^{\text{cmp}}, P^{\text{sec}}, R\}\footnote{We only consider some key parameters related to communication and computing here.}$, where $S$ is the set of services that $a_m$ can provide, $C^{\text{rate}}$ represent the average transmission rate from $a_1$ to $a_m$, $C^{\text{sec}}$ represents the security level of communication link, $P^{\text{cmp}}$ denotes the available computing power, $P^{\text{sec}}$ is the security level of computing environment, and  $R$ represents the loyalty of $a_m$ in delivering effective results. 


\begin{figure*}[!]
\centering
\includegraphics[scale=1]{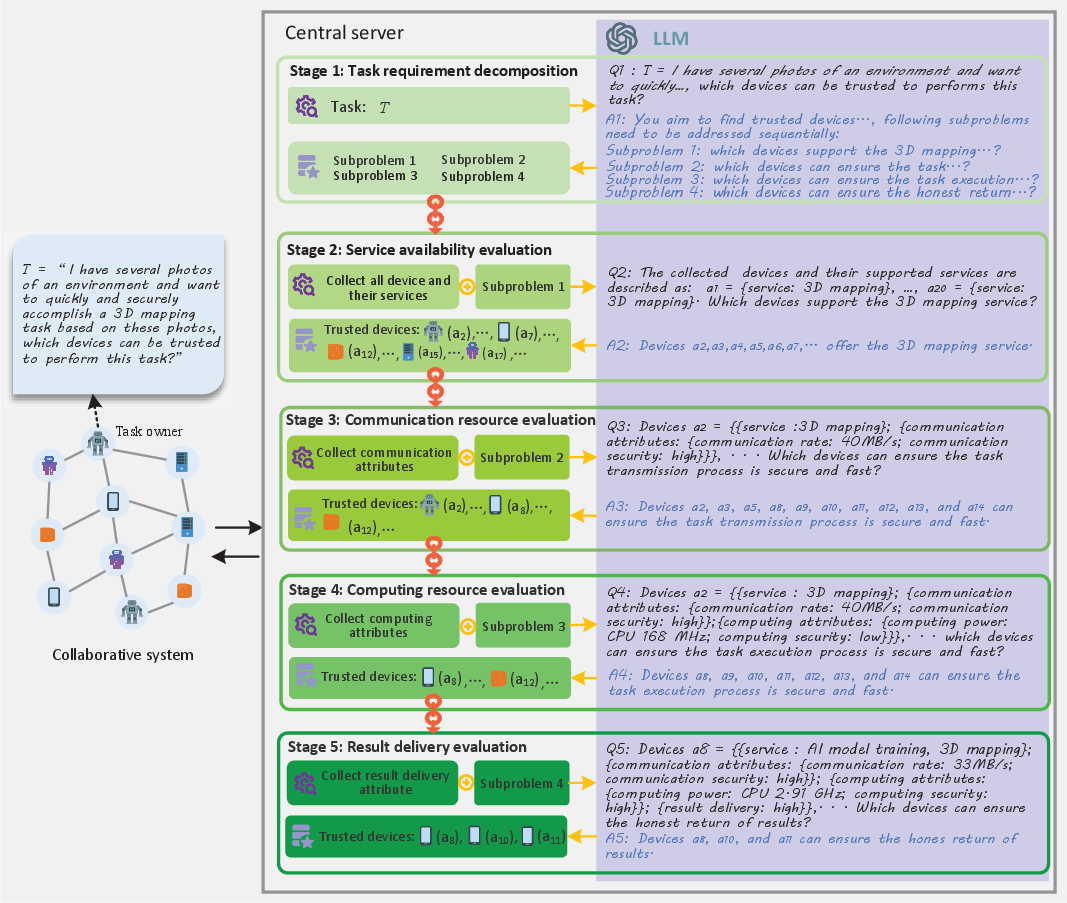}
\caption{Generative AI-enabled chain-of-trust.}
\label{flowchart}
\end{figure*}

\subsection{Implementation of chain-of-trust}
In this subsection, we analyze each stage of the chain-of-trust in detail, and then provide an example to explain the specific implementation of each stage through generative AI.
When dealing with problems that require complex logical reasoning, LLMs may struggle to provide accurate answers in a single response. To mitigate the impact of this limitation on trust evaluation, we decompose the trust evaluation problem into simpler subproblems. Each subproblem corresponds to a phase in the chain-of-trust.
These subproblems are solved sequentially, with each answer depending on the solution to the previous one. Each step is implemented through few-shot chain-of-thought prompting, eliminating the need for training or fine-tuning.



\textbf{Stage 1: Task requirement decomposition}

\textit{1.1) Analysis}: Different tasks have varying service and resource requirements, and potential collaborators must meet these requirements. For instance, an AI model training task requires collaborators to provide AI model training services while having high-performance GPU and energy resources. These services and resources are typically beyond the capacity of ordinary devices. Therefore, the first step of chain-of-trust is to conduct an in-depth analysis of tasks from task owner to identify the key services and resources required, including computing, communication, storage, networking, energy, and security, to name a few. Based on the task analysis, the trust evaluation problem is decomposed into several subproblems.

\textit{1.2) Implementation}: The task owner $a_1$ sends the task $T$ to the central server, which then initiates the execution of chain-of-trust. The central server analyzes the services and resources required for the task and decomposes the trust evaluation problem into several subproblems. The few-shot chain-of-thought prompting method is used to teach the LLM how to analyze the task by providing four exemplars. An example of such an exemplar is provided below:

Exemplar: ``\textit{I would like to complete a video capture task quickly and securely, which devices can be trusted to perform this task?}''

``\textit{You aim to find trusted devices that can complete a video capture task quickly and securely. To achieve this, the following subproblems need to be addressed sequentially:
subproblem 1: which devices support the video capture service?
subproblem 2: which devices can ensure the task transmission process is secure and fast?
subproblem 3: which devices can ensure the task execution process is secure and fast?
subproblem 4: which devices can ensure the honest return of results?}''

Then, we input the task $T$ into the LLM and obtain the following response: 

Q1: ``$T$ = \textit{I have several photos of an environment and want to quickly and securely accomplish a 3D mapping task based on these photos, which devices can be trusted to perform this task?}''

A1: ``\textit{You aim to find trusted devices that can complete a 3D mapping task quickly and securely. To achieve this, the following subproblems need to be addressed sequentially: subproblem 1: which devices support the 3D mapping service? subproblem 2: which devices can ensure the task transmission process is secure and fast? subproblem 3: which devices can ensure the task execution process is secure and fast? subproblem 4: which devices can ensure the honest return of results?}'' 


\textbf{Stage 2: Service availability evaluation}

\textit{2.1) Analysis}: Due to differences in hardware and software resources, different devices are capable of supporting different services. At this stage,  the central server collects and organizes the service information that each device can provide and matches it with the task's service requirements. Only devices that can support the required services will proceed to the next stage of trust evaluation.

\textit{2.2) Implementation}: This stage focuses on solving subproblem 1. The central server gathers all devices and their supported services within the system. The gathered data is then fed into the LLM, which filters out devices based on the task's service requirements. Few-shot chain-of-thought prompting is applied to teach the LLM using four exemplars. Each exemplar follows the format $(\text{question}, \text{answer})$, with one example presented below.

Exemplar: ``\textit{The collected devices and their supported services are described as: $a_1$ = \{\{service : 3D mapping\}\}, \dots,  $a_{20}$ = \{\{service : 3D mapping\}\}. $a_1$ is the task owner. Which devices support the AI model training service?}''

``\textit{Devices $a_7$, $a_8$, $a_9$, $a_{10}$, $a_{11}$, $a_{12}$, $a_{13}$, $a_{14}$, $a_{15}$, and $a_{16}$ offer the AI model training service.}'' 

Then, we input the collected data and subproblem 1 into the LLM.

Q2 (Devices and their services + subproblem 1): ``\textit{The collected devices and their supported services are described as: $a_1$ = \{\{service : 3D mapping\}\}, \dots,  $a_{20}$ = \{\{service : 3D mapping\}\}. $a_1$ is the task owner. Which devices support the 3D mapping service?}''

A2: ``\textit{Devices $a_2$, $a_3$, $a_4$, $a_5$, $a_6$, $a_7$, $a_8$, $a_9$, $a_{10}$, $a_{11}$, $a_{12}$, $a_{13}$, $a_{14}$, $a_{17}$, $a_{18}$, $a_{19}$, $a_{20}$ offer the 3D mapping service.}''


\textbf{Stage 3: Communication resource evaluation}

\textit{3.1) Analysis}:
Based on the devices selected in the previous round, we continue to evaluate their communication resources for effective task transmission. The evaluation targets two critical dimensions: communication security and successful task delivery. In terms of communication security, it is crucial to evaluate whether potential collaborators can provide reliable security mechanisms to ensure the confidentiality and integrity of the communication process, such as hardware-level security protections, identity authentication, data encryption, and secure communication protocols. For successful task delivery, the evaluation should determine whether potential collaborators possess the resources and mechanisms required for communication, such as sufficient bandwidth, spectrum resources, and data retransmission mechanisms. Therefore, the central server collects the aforementioned data and evaluates the devices' trustworthiness in task transmission accordingly. Devices deemed trustworthy in this stage will proceed to the next round of evaluation.

\textit{3.2) Implementation}: This stage is dedicated to addressing subproblem 2. The central server collects the values of $C^{\text{rate}}$ and $C^{\text{sec}}$ for all devices selected in the previous round. Since $a_1$ connects to other devices via WIFI that uses AES encryption, the value of $C^{\text{sec}}$ for all devices is set to a high security level. Similarly, the few-shot chain-of-thought prompting method is first utilized. Next, the prompt is created by appending subproblem 2 to devices with the updated communication attributes and then input into the LLM.

Q3 (A2 with updated communication attributes + subproblem 2): ``\textit{Devices $a_2$ = \{\{service : 3D mapping\}; \{communication attributes: \{communication rate: 40MB/s; communication security: high\}\}\},\dots, $a_{20}$ = \{\{service : 3D mapping\}; \{communication attributes: \{communication rate: 1MB/s; communication security: high\}\}\}, which devices can ensure the task transmission process is secure and fast?}''

A3: ``\textit{Devices $a_2$, $a_3$, $a_5$, $a_8$, $a_9$, $a_{10}$, $a_{11}$, $a_{12}$, $a_{13}$, and $a_{14}$ can  ensure the task transmission process is secure and fast.}''


\textbf{Stage 4: Computing resource evaluation} 

\textit{4.1) Analysis}: Based on the previous evaluation results, the computing resources of the selected devices are further assessed to ensure effective task execution. This assessment encompasses two principal dimensions: successful task execution and data security. Regarding successful task execution, it is necessary to assess the collaborators' computing-related resources and performance, such as CPU, GPU, RAM, and throughput, to determine whether these resources can ensure successful task completion. In terms of data security, it is essential to evaluate whether collaborators have mechanisms in place to ensure data protection, such as Docker-based task isolation, data encryption, and trusted execution environments. Therefore, the central server gathers the aforementioned data and evaluates the devices' trustworthiness in task execution. Devices that meet the trust criteria will proceed to the next round of trust evaluation.

\textit{4.2) Implementation}: This stage addresses subproblem 3 by evaluating the trust of devices in task execution. The central server gathers the values of $P^{\text{cmp}}$ and $P^{\text{sec}}$ for all devices selected in the previous round. Since Rosbot Plus and Robofleet robots operate on the open-source ROS system, their computing environments are classified as having a low security level. To guide the LLM, four exemplars are initially provided. The prompt is crafted by combining devices that have updated computing attributes with subproblem 3, and then it is inputted into the LLM.

Q4 (A3 with updated computing attributes + subproblem 3): ``\textit{Devices $a_2$ = \{\{service : 3D mapping\}; \{communication attributes: \{communication rate: 40MB/s; communication security: high\}\};\{computing attributes: \{computing power: CPU 168 MHz; computing security: low\}\}\},\dots, $a_{14}$ = \{\{service : 3D mapping, edge computing\}; \{communication attributes: \{communication rate: 35MB/s; communication security: high\}\};\{computing attributes: \{computing power: CPU 3.8 GHz; computing security: high\}\}\}, which devices can ensure the task execution process is secure and fast?}''

A4: ``\textit{Devices $a_8$, $a_9$, $a_{10}$, $a_{11}$, $a_{12}$, $a_{13}$, and $a_{14}$ can ensure the task execution process is secure and fast.}''


\textbf{Stage 5: Result delivery evaluation} 

\textit{5.1) Analysis}: Based on the devices selected in the previous step, the server continues to collect data related to result delivery to assess their trustworthiness in delivering results. The collaborators selected in this step are the trustworthy devices that fully meet the task requirements.

\textit{5.2) Implementation}: This step addresses subproblem 4 by evaluating the trustworthiness of devices in result delivery. The central server gathers the result delivery behaviors of all devices selected in the previous round. The prompt is constructed by merging devices with updated result delivery attribute and subproblem 4, then inputted into the LLM.

Q5 (A4 with updated result delivery attribute + subproblem 4): ``\textit{Devices $a_8$ = \{\{service : AI model training, 3D mapping\}; \{communication attributes: \{communication rate: 33MB/s; communication security: high\}\};\{computing attributes: \{computing power: CPU 2.91 GHz; computing security: high\}\}; \{result delivery: high\}\},\dots, $a_{14}$ = \{\{service : 3D mapping, edge computing\}; \{communication attributes: \{communication rate: 35MB/s; communication security: high\}\};\{computing attributes: \{computing power: CPU 3.8 GHz; computing security: high\}\}; \{result delivery: low\}\}, which devices can ensure the honest return of results?}''

A5: ``\textit{Devices $a_8$, $a_{10}$, and $a_{11}$ can ensure the honest return of results.}''

Finally,  $a_8$, $a_{10}$, and $a_{11}$ selected at this stage are the trusted collaborators capable of effectively completing task $T$.

\subsection{Dynamism and flexibility of chain-of-trust}
Chain-of-trust is both dynamic and flexible, characterized by two main features. First, the number of evaluation stages is task-specific and can be adjusted in response to changing task requirements. Second, each stage supports repeated execution, enabling periodic reassessment of device trust. Such adaptability ensures accurate and task-aligned trust evaluation in dynamic and heterogeneous environments.}



\section{Simulation and Performance Evaluation}

In this subsection, extensive experiments are conducted to validate the trust evaluation accuracy of the proposed chain-of-trust.  $a_1$ is assumed to be the task owner, generating 200 tasks, with the parameters of all collaborators as shown in Table I. The correct answers for these tasks are obtained via manual reasoning.

\begin{figure}[t!]
\centering
\includegraphics[scale=0.6]{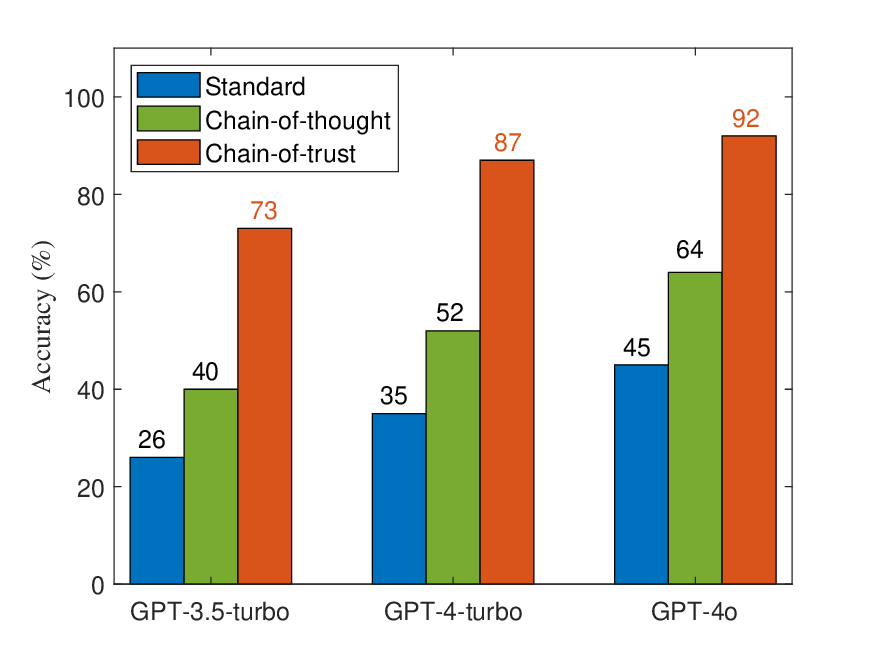}
\caption{Comparison of trust evaluation accuracy on different GPT models.}
\label{result1}
\end{figure}

The proposed chain-of-trust is tested on GPT-3.5-turbo, GPT-4-turbo, and GPT-4o models. To enable a more comprehensive evaluation, chain-of-thought and standard models are used as baseline comparisons, highlighting the advantages of the proposed approach.
As shown in Fig.~\ref{result1}, the standard GPT-3.5-turbo achieves the lowest accuracy due to its limited capability in processing and understanding long texts. The chain-of-thought approach improves accuracy to some extent, achieving accuracies of 40\%, 52\%, and 64\% on GPT-3.5-turbo, GPT-4-turbo, and GPT-4o, respectively. {While the chain-of-thought approach enhances the performance of LLMs on multi-step reasoning tasks, its effectiveness is limited by the reliance on prompt design, the potential misalignment between generated reasoning and actual decision-making, and the need for large-scale LLMs to achieve optimal results.} The proposed chain-of-trust demonstrates significant performance improvements compared to both the standard model and chain-of-thought, achieving accuracies of 73\%, 87\%, and 92\% on GPT-3.5-turbo, GPT-4-turbo, and GPT-4o, respectively. As we can see, GPT-4o consistently achieves better results due to its stronger ability to understand complex texts.

\section{Conclusions}


In this paper, we have proposed a novel progressive trust evaluation framework: chain-of-trust, for collaborative systems with limited network resources and asynchronously updated device attribute data. To implement this framework, we have seamlessly integrated generative AI into chain-of-trust, leveraging few-shot prompting to enable dynamic adaptation to new tasks without additional training. Compared to the standard GPT models and the chain-of-thought approach, the proposed framework can achieve superior accuracy in trust evaluation. Chain-of-trust provides a robust and adaptable solution that enhances the reliability and accuracy of trust evaluation in dynamic and resource-constrained environments.

\footnotesize

\end{document}